\RequirePackage{fix-cm}
\documentclass[twocolumn]{svjour3}          
\smartqed  
\usepackage{graphicx}
\usepackage{tabularx} 
\usepackage{amsmath}  
\DeclareMathOperator*{\argmax}{arg\,max}
\usepackage{natbib}
\usepackage{amsfonts}
\usepackage{mathtools}
\usepackage{amsmath}
\usepackage{amssymb}
\usepackage{framed}
\usepackage{subfigure}
\usepackage{graphics} 
\usepackage{lipsum}
\usepackage{graphicx}
\usepackage{mathptmx} 
\usepackage{gensymb}
\usepackage{algorithm}
\usepackage{amssymb}  
\usepackage{algpseudocode}
\usepackage{booktabs}
\usepackage{tabularx}
\usepackage{booktabs}
\usepackage{array}
\usepackage{etex}
\usepackage{adjustbox}
\usepackage{adjustbox,lipsum}
\usepackage[table]{xcolor}
\usepackage{nicefrac}
\usepackage{bbm}
\usepackage{mdwtab}
\usepackage[title]{appendix}
\usepackage{multirow}

\begin{document}
\sloppy

\title{Active Object Tracking using Context Estimation: \\
 Handling Occlusions and Detecting Missing Targets
}


\author{Minkyu Kim \and Luis Sentis}


\institute{Minkyu Kim \at
              2617 Wichita St, Austin, TX 78712 \\
              Tel.: +1-714-222-2011\\
              \email{steveminq@utexas.edu}           
           \and
           Luis Sentis \at
              2617 Wichita St, Austin, TX 78712 \\
              \email{lsentis@austin.utexas.edu} 
}

\date{Received: date / Accepted: date}
\maketitle

\begin{abstract}
When performing visual servoing or object tracking tasks, active sensor planning is essential to keep targets in sight or to relocate them when missing. In particular, when dealing with a known target missing from the sensor's field of view, we propose using prior knowledge related to contextual information to estimate its possible location. To this end, this study proposes a Dynamic Bayesian Network that uses contextual information to effectively search for targets. Monte Carlo particle filtering is employed to approximate the posterior probability of the target's state, from which uncertainty is defined. We define the robot's utility function via information theoretic formalism as seeking the optimal action which reduces uncertainty of a task, prompting robot agents to investigate the location where the target most likely might exist. Using a context state model, we design the agent's high-level decision framework using a Partially-Observable Markov Decision Process. Based on the estimated belief state of the context via sequential observations, the robot's navigation actions are determined to conduct exploratory and detection tasks. By using this multi-modal context model, our agent can effectively handle basic dynamic events, such as obstruction of targets or their absence from the field of view. We implement and demonstrate these capabilities on a mobile robot in real-time.
\end{abstract}

\section{Introduction}
\label{intro}
This paper addresses visual-based active object tracking using mobile sensor platforms. Recently, low cost vision sensors and object detection algorithms have been broadly available \citep{redmon2016you}, upping the development of new object tracking capabilities. Visual information of objects can easily incorporate semantic information of known targets to resolve ambiguity related to data association in cluttered environments \citep{makris2011hierarchical}. However, vision-based object tracking suffers from occlusion from overlapping objects, or missing targets from the field of view (FOV) \citep{xiang2015learning, yun2017action}. To overcome these difficulties, many researchers have investigated ``active" approaches to change or move camera states in order to keep tracking targets in sight. In other words, the focus of previous research has shifted from passive to active perception (sensing) to handle more dynamic tasks such as pick/place, and dynamic target grasping \citep{kaelbling2012unifying, eidenberger2009integrated, arora2020mobile}. However, there is little work to solve the more challenging situations involving long-term occlusions or losing sight of targets unexpectedly.  Recently, deep reinforcement learning methods have been applied for active object tracking \citep{luo2019end} and visual-servoing tasks \citep{shi2018adaptive} showing target recovery capabilities, but these methods are limited to simulation environments or do not perform active search when facing occlusions like we do in our work.

The authors of \citep{bajcsy1988active} firstly described active perception as ``a problem of controlling strategies applied to the data acquisition process which depends on the current state of the data interpretation and the task of the process". Similarly, this paper proposes an active perception framework that seeks optimal control inputs that reduce target uncertainty based on information theoretic costs \citep{eidenberger2009probabilistic}. Information theoretic approaches have been widely used in state-estimation and control of mobile sensor systems such as mapping \citep{charrow2015information, julian2014mutual}, Simultaneous localization and mapping (SLAM) \citep{valencia2018active}, or object pose estimation \citep{wu2015active}. Related to our work, \citep{ryan2010particle} uses Bayesian estimation and information theoretic cost to reduce uncertainty of target locations. 

Object tracking with Bayesian estimation lacks support for dealing with the absence of targets since direct measurements might not be available. In such situations, we convert the tracking problem into an object search problem. Object search can be solved by finding the most informative actions to re-locate lost targets. Since there is no direct observation of an object, we aim to obtain an optimal action policy based on the probabilistic belief of possible target locations. 

\citep{bourgault2003optimal} investigated optimal search strategies to minimize the expected time to find lost targets within probabilistic frameworks. Other probabilistic formulations for object search have been considered such as \citep{bertuccelli2006search, lau2006probabilistic, chung2012analysis}. Most of these methods reduce the search space using a discrete-grid world or graphical structure, since computing the optimal search actions with uncertainty is known to be an NP-hard problem \citep{tseng2017near}. Reformulating optimal search by reducing the search space mitigates the computational burden. However, these methods do not deal with dynamic situations such as occlusions or missing targets. Other works have shown progress recovering missing targets such as \citep{radmard2017active} or \citep{radmard2018resolving}. However, since these studies do not utilize semantic information of objects and only handle occlusions using geometric information, there are less versatile than our method; for instance they cannot reason whether an object is occluded or has suddenly disappeared. In addition, although an occlusion-aware planning strategy for multiple robots has been studied using an information theoretic approach \citep{hausman2016occlusion}, this work has the critical assumption that tracking never fails completely because of occlusions. Recently, Shi \citep{shi2018adaptive} applied reinforcement learning to resolve occlusions during image-based visual servoing tasks, focusing on finding missing features from QR code images without using semantic information or context. Once again, such method provides less versatility to complex situations than our method.

\begin{figure}[t]
    \centering
        {\includegraphics[width=1.0\linewidth]{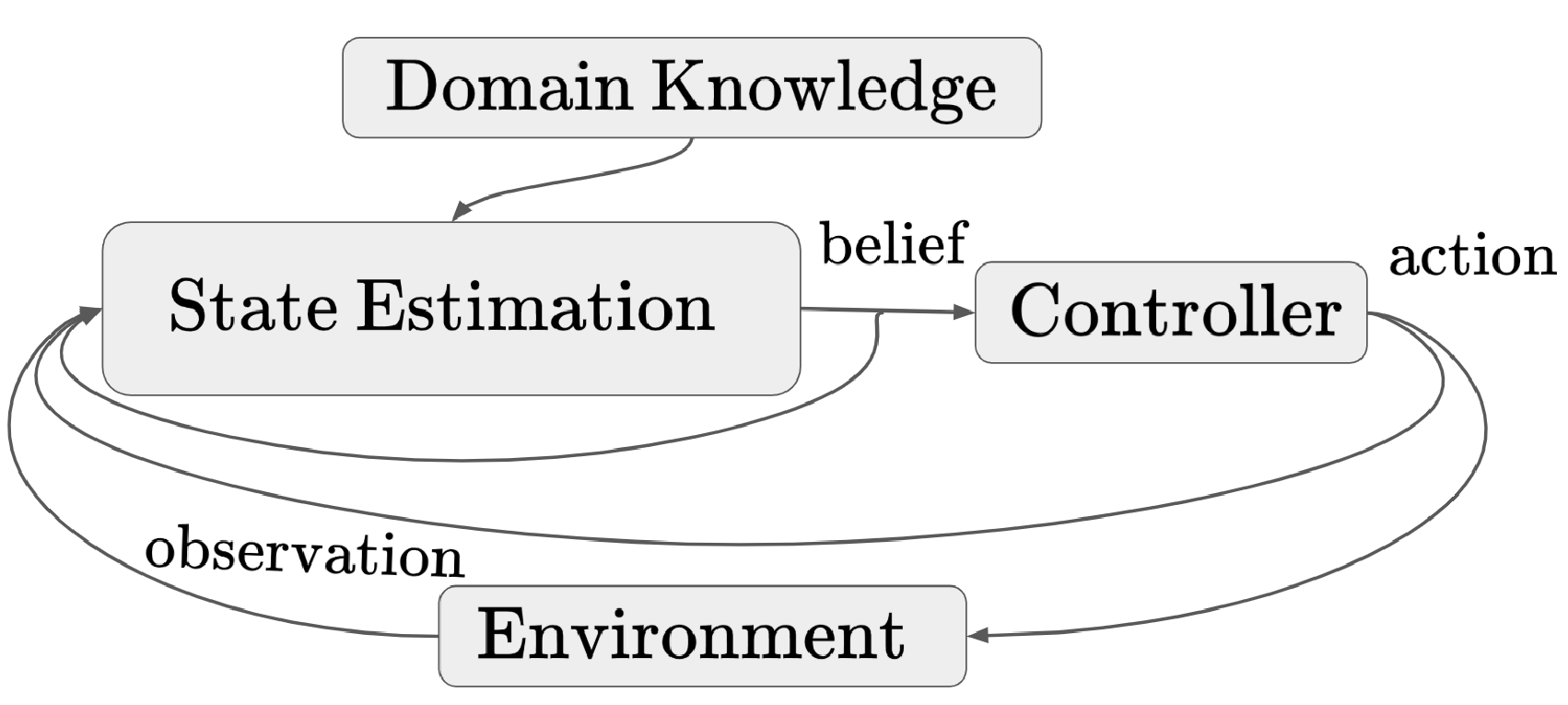}}
       \caption{ }
       \label{Figure:Concept-Diagram}
\end{figure}

Inferring the possible location of targets draws inspiration from human perceptual capabilities \citep{aydemir2013active}. Our research proposes estimating possible target locations based on predicted context information. By using prior knowledge of the target's surrounding or past experience of the sensing process, the quality of a target's prediction can be improved significantly. From this perspective, providing high-level information (domain knowledge or context) to a state estimator is strongly desirable \citep{denzler2002information, kaelbling2013integrated}. The main challenge with this concept is that this estimation framework needs to be formulated so that it can be used simultaneously to address tracking, searching of objects, and recovering from occlusions. To resolve this difficulty, our study proposes to utilize domain knowledge for target estimation using probabilistic tools such as Dynamic Bayesian Networks (DBN), particle filters, and POMDP. 

We estimate context states using probabilistic beliefs. Computing an optimal action over the belief space can be formulated as a Partially-Observable Markov Decision Process (POMDP). There exist approximate solutions for low dimensional systems \citep{porta2006point} and locally optimal solutions for continuous state and action spaces \citep{van2012motion, bai2014integrated}. Furthermore, online POMDP solvers for quite large dimensions are studied. \citep{silver2010monte, ye2017despot}. Similarly to \citep{sridharan2010planning, li2016act}, we utilize POMDP to high-level planning in order to achieve real-time performance with an online solver. 

By estimating their context, our robots can take optimal decisions based on increasing the information gain. For example, when a target object is present, our robot will attempt to keep targets in its FOV by controlling its head or mobile base. When an occlusion occurs, our robot will move to the optimal sensor configuration obtained by solving the alluded information theoretical control problem. In addition, when targets suddenly disappear, our robot will search for them based on context information. For our study, we assume that objects can not move by themselves but only by nearby people. With this assumption, our robot's logical action is to find targets near observed people. When no person is present, our robot will start exploring the area around it to locate people.
  
In summary, the main contributions of this work are 1) devising a probabilistic framework for leveraging context, which adds a situational layer of information for more versatile active object tracking (see Fig.~\ref{Figure:POMDP-STATES} for an example of situational context states) such as resolving occlusions and missing, 2) building a bridge between particle filtering and POMDP enables achieving not only high-level decision-making but also constructing a sampling-based optimization problem to find the optimal sensor configuration using the information theoretic approach. Lastly, 3) integrating a realistic demonstration with a mobile robot for validating the newly defined capabilities.  

\section{State Estimation}
\begin{table}[t]
\caption{Nomenclature}
\centering
\begin{tabular}{cc}
\hline
\hline
Symbol & description \\
\hline
$x$    & target states \\
$q$  & robot (sensor) state configurations  \\
$u$  & robot inputs \\
$c$  & context states \\
$z$  & measurement of target states\\ 
$\Theta$  & measurement of context states \\
\hline
\hline
\end{tabular}
\label{table_parameter}
\end{table}

\subsection{Bayesian Filtering: Active Object Tracking}
Bayesian filtering is a great tool to estimate the state of dynamic systems recursively in a probabilistic manner. This approach attempts to construct the posterior probability of a state based on a sequence of observations. In general, an active object tracking problem (object tracking with active sensing) can be formulated with a set of models: motion transition model $p(x_{k}|x_{k-1})$, sensor (robot) motion model $p(q_{k}|q_{k-1},u_{k-1})$, and measurement model $p(z_k|x_k,q_k)$, where $x_k \in \Re^3$, $q_k$, $u_k$, and $z_k$ are target states, sensor (robot) states, control inputs, and observations at time step $k$, respectively. In this study, a target state is described as the 3D position of an object, and $q_k$ denotes the robot's base and its camera's configuration, namely, $q_k = (x,y,\theta, q_{tilt},q_{pan})$. Here, a sensor motion model is the probabilistic form of the robot's forward dynamics, which assumes that $q_{k+1}$ is observable. In addition, $z_k$ is the detection output through an object recognition algorithm and point cloud processing, encoding the 3D position of the observed targets.

The goal of filtering is to estimate the target state $x$ at time $k$, namely, the posterior distribution, using the a priori estimate (prediction step) and the current measurement of the sensor (update step). Assuming that the prior probability $p(x_{k-1}|z_{1:k-1})$ is available at time $k-1$, the prediction step attempts to estimate $P(x_k|z_{1:k-1})$ from previous observations as follows:
\begin{equation}
    p(x_k|z_{1:k-1})=\int p(x_k|x_{k-1})p(x_{k-1}|z_{1:k-1})dx_{k-1},
\label{Eq:1}
\end{equation}
where $p(x_k|x_{k-1})$ is the target's motion model based on a first order Markov process. Then, when the measurement $z_k$ is available, given sensor configuration $q_k$ at time $k$, the estimated state can be updated as 
\begin{equation}
    p(x_k|z_{1:k})=\frac{p(z_k|x_k,q_k)p(x_k|z_{1:k-1})}{p(z_k|z_{1:k-1},q_k)},
\label{Eq:2}
\end{equation}
where $p(z_k|z_{1:k-1},q_k)=\int p(z_k|x_k,q_k)p(x_k|z_{k-1})dx_k$. For the update step, the measurement $z_k$ is used to modify the prior estimate, leading to obtaining the posterior distribution of the current state. Finally, after calculating the target's belief state $p(x_k | z_k)$, the goal of active sensing is to provide optimal sensor control inputs to track the target or to find it if its lost. 

\subsection{Context Modeling}
In general, for object tracking, a transition model is assumed to be known based on a constant velocity model hypothesis with white noise, i.e. $P(x_{t}|x_{k-1})= x_{k-1}+v_{k-1}\Delta t+\nu$ where $\nu$ represents white noise. However, this approach is inefficient in active object tracking since targets might be frequently missing, meaning that $v_k$ is unavailable. Thus, inspired by how humans seem to track objects, it would be beneficial to understand the current situational context in order to effectively estimate possible locations of lost targets. To leverage this abstract knowledge or context, we employ a Dynamic Bayesian Network model to infer the possible state of targets. Consequently, when an agent begins looking for a missing target it will benefit by knowing the belief state of the context and trying to find it based on such information. 

\subsubsection{Dynamic Bayesian Networks}
\begin{figure}
    \centering
        {\includegraphics[width=0.75\linewidth]{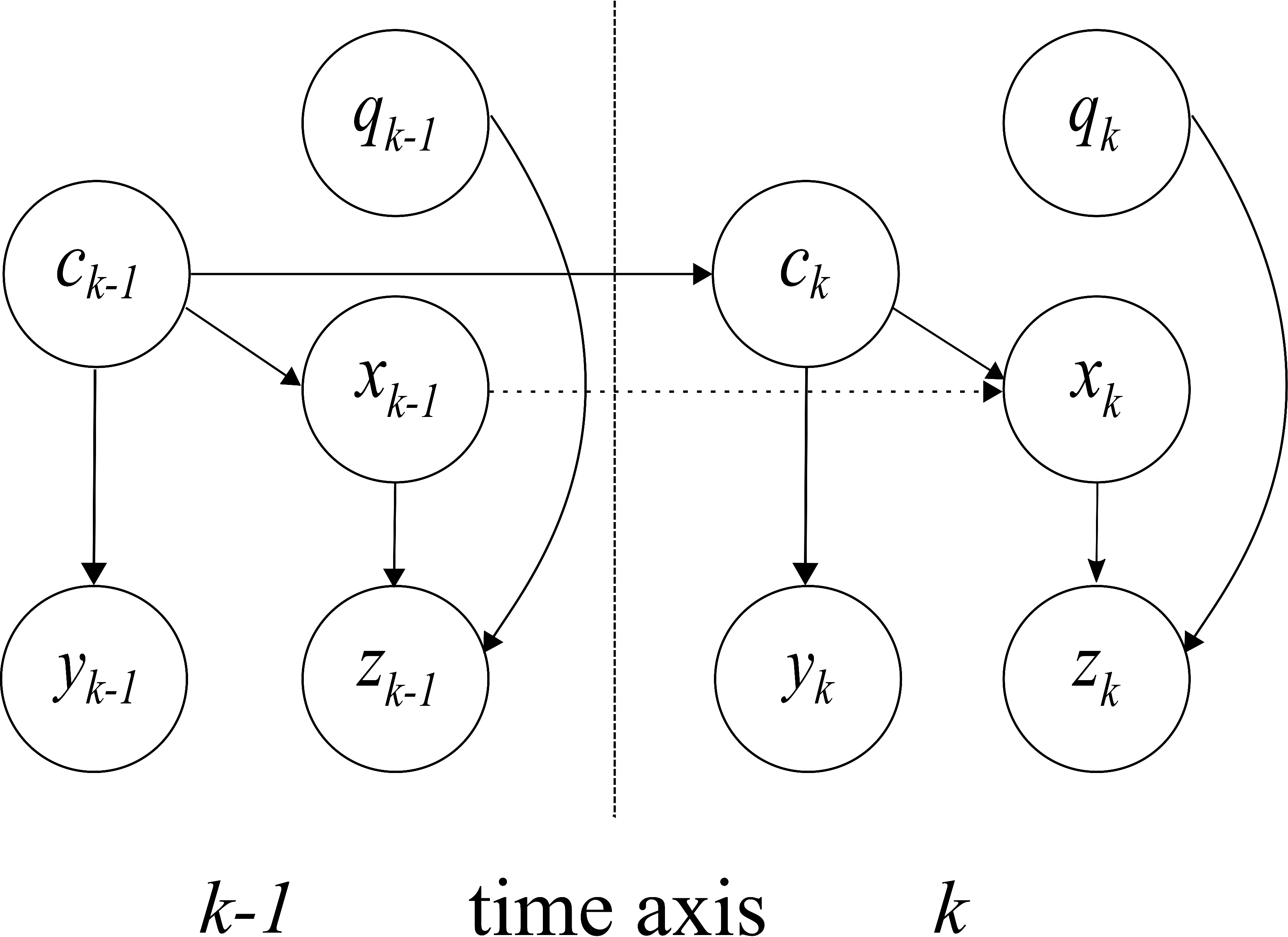}}
       \caption{Dynamic Bayesian Network with context states ($c_k$), target states ($x_k$), and robot configurations ($q_k$). $y_k$ and $z_k$ are measurements for context and targets, respectively. The arrows indicate the conditional dependencies between the random variables. The dotted line indicates that when an object is missing, it's target location cannot be directly estimated. In our case, it is inferred from context information.}
       \label{Figure:Context-Bayesian-Network}
\end{figure}

DBN is an extension of Bayesian Networks, a graph model for representing causal relations and conditional dependencies, for  temporal processes, which can evolve over time. The proposed model is shown in Fig. \ref{Figure:Context-Bayesian-Network}, where $c_k$ denotes the context state at time step $k$, and $y_k$ is the observation of the context state. Since target states can depend on context states, this model can be directly applied to Bayesian filtering as a transition model for  prediction. In other words, instead of using $p(x_{k}|x_{k-1})$, we propose to use $p(x_{k}|c_{k-1})$ as the motion model, i.e.
\begin{equation}
    p(x_k|z_{1:k-1})\approx\int p(x_k|c_{k-1})p(x_{k-1}|z_{1:k-1})dx_{k-1},
\label{Eq:neww}
\end{equation}
In this sense, target states can be predicted using context states, which in turn can be estimated using context transition models and context measurements.

\subsubsection{Hybrid Motion Model}
More precisely, the target transition model can be reformulated using context states as 
\begin{equation}
p(x_{k}|c_{k-1}) = \int p(x_{k}|c_{k}) p(c_{k}|c_{k-1})dc_{k}. 
\label{motion_equation}
\end{equation}
Here, we assume that $p(c_{k}|c_{k-1})$ is known (this will be explained in section \ref{pomdp_transition}). Furthermore, if the set of possible context states is finite, the posterior distribution of $x_k$ can be calculated as 
\begin{equation}
p(x_{k}|c_{k-1}) = \sum^N_{i=1}  p(x_{k}|c^i_{k})p(c^i_{k}|c_{k-1}),
\label{eq:xkck-1}
\end{equation}
where $c^i_k$ denotes the $i$-th context state at time $k$ and $N$ is the cardinality of the set of context states. If each context $c_k$ is described using a Gaussian model, the posterior distribution becomes a Gaussian Mixture Model (GMM). As such, the probability density of the GMM is equivalent to the weighted sum of all components, i.e. 
\begin{equation}
p(x_{k}|c_{k-1})= \sum^N_{i=1} p(c^i_{k})\mathcal{N}(x_{k};\mu_i,\Sigma_{i}),
\label{Equation:GMM}
\end{equation}
where $\mathcal{N}$ is the multivariate Gaussian model with mean vector $\mu$ and co-variance matrix $\Sigma$. Here, each $\mu_i$ and $\Sigma_i$ represent the mean target location given the $i$-th context and its variance. 

\subsection{Particle Filter}

We approximate the posterior distribution $p(x_k|z_{1:k})$ by a set of $N$ number of particles $\{x_k,w_k\}^N$, where $w_k$ is the corresponding weight of the particle $x_k$ at time step $k$. Generally, the weighted approximation is calculated as 
\begin{equation}\label{eq:pparticles}
    p(x_k|z_{1:k})\approx \sum^N_{i=1}w_k^i\delta(x_k-x^i_k),
\end{equation}
where particles $x^i_k$ are drawn from a proposal density $q(x_{0:k}|z_{1:k})$ and the weight of the particles is updated using importance sampling. The normalized weight of the $i$-th particle can be defined as
\begin{equation}
    w^i_k=\frac{p(x^i_{0:k}|z_{1:k})}{q(x^i_{0:k}|z_{1:k})}.
\end{equation}

Assuming the Markov property, the proposal density can be decomposed into the prior proposal density and the propagated density as 
\begin{equation}
q(x_{0:k}|z_{1:k})=q(x_{k}|x_{k-1},z_{1:k})q(x_{0:k-1}|z_{1:k-1}).
\end{equation}
Thus, this equality yields  
\begin{eqnarray}
  w^i_k&=& \frac{p(z_k|x_k^i)p(x_k^i|x_{k-1})p(x^i_{0:k-1}|z_{1:k-1})}{q(x^i_{k}|x^i_{k-1},z_{1:k})q(x^i_{0:k-1}|z_{1:k-1})} \\
      &=& \frac{p(z_k|x_k^i)p(x_k^i|x_{k-1})}{q(x^i_{k}|x^i_{k-1},z_{1:k})}w^i_{k-1}
\end{eqnarray}
Here, a common choice for $q(x^i_k | x^i_{0:k-1},z_{1:k})$ is the motion model $p(x_{k}|x_{k-1})$, which minimizes the variance of $p(x|z)$. Also, as mentioned before, in the above equations we use Eq. \eqref{eq:xkck-1} to estimate the process $p(x_k^i|x_{k-1})$. Based on Bayesian filtering, at every time step $k$, these weights can be recursively updated from the time step $k-1$, 
\begin{equation}
    w^i_k=w^i_{k-1}\frac{p(z_k|x_k^i,q_k)}{\sum^N_{j=1}p(z_k|x^j_k, q_k)}.
\end{equation}

\subsection{Sensor Model}
\label{sensor_model}
The sensor model, $p(x|z,q)$, has the form of a joint distribution for $z$, $x$, and $q$. This model must consider the case of an empty measurement set (i.e. when the target is missing). In addition, the sensor model depends on whether the target is within the FOV or not. The FOV can be formulated as a function of $q$, $f(q)$, given the robot's configuration \citep{freda2008sensor}. For an RGB-D camera, $f(q)$ has a cone shape with center $c(q)$, opening angle $\alpha$, and radius $R$. These parameters are later determined based on real-world sensor specifications. In this study, the sensor model is defined as
\begin{align}
\label{Eq:Sensor_model}
  \begin{split}
   \begin{array}{lrr}
  p(z=\emptyset |x,q) &= 1-p_e \qquad \qquad \quad \; &\text{if} \; x \not\subset f(q)\\
  p(z=\emptyset |x,q) &= 1-p_d \qquad \qquad \quad \; & \text{if} \; x \subset f(q) \\
  p(z \neq \emptyset |x,q) &= p_e \qquad \qquad \qquad  \;\; \; \; &  \text{if} \; x \not\subset f(q) \\
  p(z \neq \emptyset |x,q) &= p_d \mathcal{N}(z;x,\Sigma^2) \qquad & \text{if} \; x \subset f(q)
 \end{array}
  \end{split}
\end{align}
where $p_d$ is a user-defined true positive probability of the detection algorithm and $p_e$ is the false negative probability ($e$ stands for error). 

\subsection{Importance Sampling}
\begin{figure}[t]
    \centering
        {\includegraphics[width=1.0\linewidth]{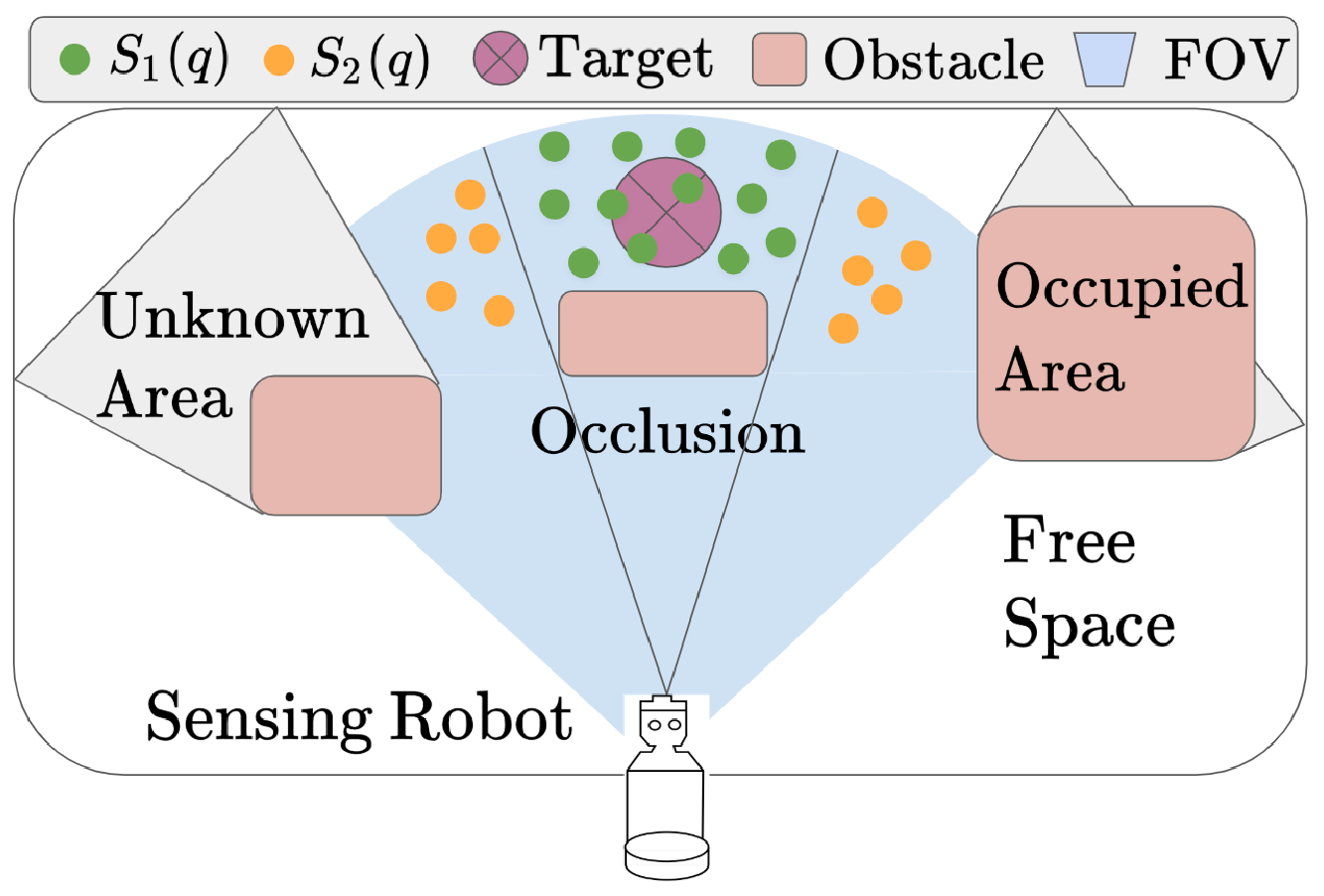}}
       \caption{ An illustration of our problem, including the robot's FOV, a target object, and its environment. The target (purple circle) is occluded by the pink object so that it cannot be detected by the robot. We separate particles into two sets, $S_1(q)$ and $S_2(q)$, based on their visibility by the robot.}
       \label{Figure:Field-of-View}
\end{figure}

Typically, sequential sensor observations are used to update weights of particles and compute the resulting probability distributions. However, for our active object tracking problem, continuous sensor readings are not always available due to frequently missing targets from occlusion or disappearance. To update the importance of particles, our proposed sensor model, Equation (13), is used. For example, for the case of target occlusion, the weight of particles in the field of view ($S_2(q)$ in Fig.~\ref{Figure:Field-of-View}) will be reduced according to the value (1-$p_d$). On the other hand, the  weights of particles outside the FOV ($S_1(q)$) will be updated according to the value (1-$p_e$). We use re-sampling to avoid degeneracy problems and increase efficiency.


\section{Methods}

\subsection{Entropy to Reduce Uncertainty}

The goal of active perception is to gather as much information as possible, or equivalently to reduce uncertainty. Shannon's entropy is defined as measuring the uncertainty of a random variable. For our problem, the desired control actions (i.e. moving the robot to a new configuration) correspond to policies that reduce uncertainty the most. Using the target's posterior distribution, $p(z|x)$, its entropy is represented as 
\begin{equation}
H(p(x|z,q))=\int_{x\in X}-p(x|z,q)logp(x|z,q)dx.
\end{equation}

Using eq.~(\ref{eq:pparticles}), the entropy can be approximated by the particle weights as
\begin{equation}
H(p(x|z,q))=-\sum_i^N w_i \log w_i
\end{equation}

The full set of particle weights $\Omega$ can be divided into two sets ($S_1(q_k)$, $S_2(q_k)$) based on the FOV of the sensor $f(q_k)$. Here $S_1$ denotes the set of particles within the FOV, while $S_2$ is the set of particles that are not included in the FOV. Note that the region defined by $S_1(q)$ does not belong to $f(q_k)$ since it is occluded. Mathematically, $S_1=\{w^i \in \Omega : x^i \notin f(q_k)\}$ and is disjointed from $S_2$, i.e. $S_2= \Omega \setminus S_1$). As a result, the entropy equation can be expressed as 
\begin{equation}
 H(p(x_k|z_k,q_k)) \approx -\sum_{i \in S_1} w_i \log w_i -\sum_{j \in S_2} w_j \log w_j.
 \label{equation:entropy_with_fov}
\end{equation}

The expected information gain (IG) only depends on the first term of the above equation, and thus can be written as
\begin{align}
\begin{split}
    IG(q_{k}) &= -E[H(x_{k-1}|z_{k-1},q_{k-1})-H(p(x_{k-1}|z_{k-1},q_{k}))] \\
     &\approx \sum_{i \in S_1(q_{k})}w_i \log w_i. \end{split}
\end{align}


\subsection{Utility Function For Target Detection}

The first term of our utility function is the information gain, which involves the amount of information that can be obtained given sampled configurations. The second term penalizes traveling cost. Lastly, we define a perception gain that assists the robot to approach targets to increase the quality of the sensing process. Mathematically, the cost function is described as
 \begin{gather}
          J(x_{k},q_{k-1},q_{k})= IG - \zeta_{travel} - \zeta_{perception} \\
          \zeta_{travel} = \beta (q_k-q_{k-1})^TQ(q_k-q_{k-1}) \\
           \zeta_{perception} = \gamma (x_{k}-S_d\, q_{k})^T(x_{k}-S_d\,q_{k}).
 \end{gather}
 Here $x_k$ is the estimated position of the target at time $k$, $q_{k-1}$ is the current robot configuration, $q_k$ are sampled robot configurations, $\beta$, $\gamma$ are cost weighting factors, and $S_d$ is a mapping that extracts the Cartesian position of the robot from its position-rotation configuration. Solving this cost yields,
 \begin{equation}
    q^*_{k} = \argmax J(x_{k},q_{k-1},q_{k})
 \end{equation}
To solve this problem we use a sampling based approach and greedy search to obtain the final result, $q_k^*$. 
 
\subsection{POMDP High-level Planner  }
\label{pomdp_chapter}
For our decision framework, we will estimate the current context and based on it choose actions to obtain rewards. Since context cannot be directly measured, our planner is formulated as a Partially-Observable Markov Decision Process (POMDP). A POMDP can be described by the tuple $(S, I, A, Z, T, R)$, a finite set of states $S=\{s_1,\cdots s_{|S|}\} $, an initial probability distribution over these states $I$, a finite set of actions $A=\{a_1,\cdots a_{|A|}\} $, a finite set of observations $Z=\{o_1,\cdots o_{|Z|}\} $, and a transition function $T(s,a,s')=P(s'|s,a)$ that maps $S\times A$ into discrete probability distributions over $S$. In detail, a transition model $T(s,a,s')$ specifies the conditional probability distribution of shifting from state $s$ to $s'$ by applying action $a$. $Z(s',a,o)=P(o|s,a)$ is the observation mapping that computes the probability of observing $o$ in state $s'$ when executing action $a$. $R=r(s',a,s)$ is the reward function.

A POMDP is equivalent to a continuous-state Markov Decision Process where the states are beliefs, also called a belief MDP \citep{krishnamurthy2016partially}. Thus a state can be rewritten using belief states $b(s)=p(s)$, defined as the posterior distribution over all possible states given the history of actions and observations. For our purposes, the states will be the context situations. In the vein of Eqs. \eqref{Eq:1} and \eqref{Eq:2}, using Bayesian formulation belief states can be represented recursively as follows:
\begin{align}
   b_{k}(s|a_{k-1})&=\sum_{s' \in S} T(s,a_{k-1},s')b_{k-1}(s) \\
   b_k(s|a_{k-1},o_k)&=\eta Z(s,a_{k-1},o_k)b_{k}(s|a_{k-1}),
 \label{Equation: Belief Updates}
\end{align}
where $\eta$ is a normalizing constant. The goal of the POMDP is to select a sequence of actions over time to maximize the expected cumulative reward. Value iteration algorithms are used for optimally solving POMDPs. This strategy can be defined as a policy $\pi^*$, which maps a belief $b$ to actions. Given policy $\pi$, a belief $b(s) \in B$, the value function can be computed via the Bellman equation as 
\begin{equation}
  V(b, \pi) = \rho(b, \pi(b))+\gamma \sum_{b' \in B} \tau(b,\pi,b') V(b',\pi),
\end{equation}
where $\gamma$ is a discount factor, $\rho$ is the expected reward, and $\tau$ is the transition probability to $b'$ from $b$ under $\pi$, which can be computed as~\citep{ross2008online}
\begin{equation}
\tau(b,\pi,b')=\sum_{o \in Z}p(b'|b,\pi,o)\; p(o|b,\pi).
\end{equation}
The best policy $\pi^{*}$ is obtained by solving the optimization problem:
\begin{equation}
\pi^{*}(b) = \argmax V(b_k, \pi_k).
\end{equation}
In this paper, we utilize an on-line POMDP solver, the Determinized Sparse Partially Observable Tree (DESPOT) \citep{ye2017despot}, to obtain the optimal action policy based on the current belief of context states. 

\subsection{POMDP Problem Formulation}
\subsubsection{System States}

 System states include robot position, target position and context states. Positions of the robot and target are represented in 2D using an occupancy grid. Context states correspond to one of the following semantically grounded symbols:$\{$\textit{Visible, Occluded, Disappearance, Irrecoverable}$\}$, some of which cannot be measured directly using sensors. \textit{Visible} is the state in which the target can be directly measured by the robot's sensors. \textit{Occluded} is the state when the target has been occluded by another object but is  believed to be behind it. \textit{Disappearance} is the state in which the target is not directly visible by the robot and is not believed to be occluded by another object in the robot's field of view. In this state, the target is believed to have been moved away by a person. Finally, \textit{Irrecoverable} corresponds to the case when the robot does not have information about the whereabouts of the target nor it can use context information to find it.
\begin{figure}[t]
    \centering
        {\includegraphics[width=1.0 \linewidth]{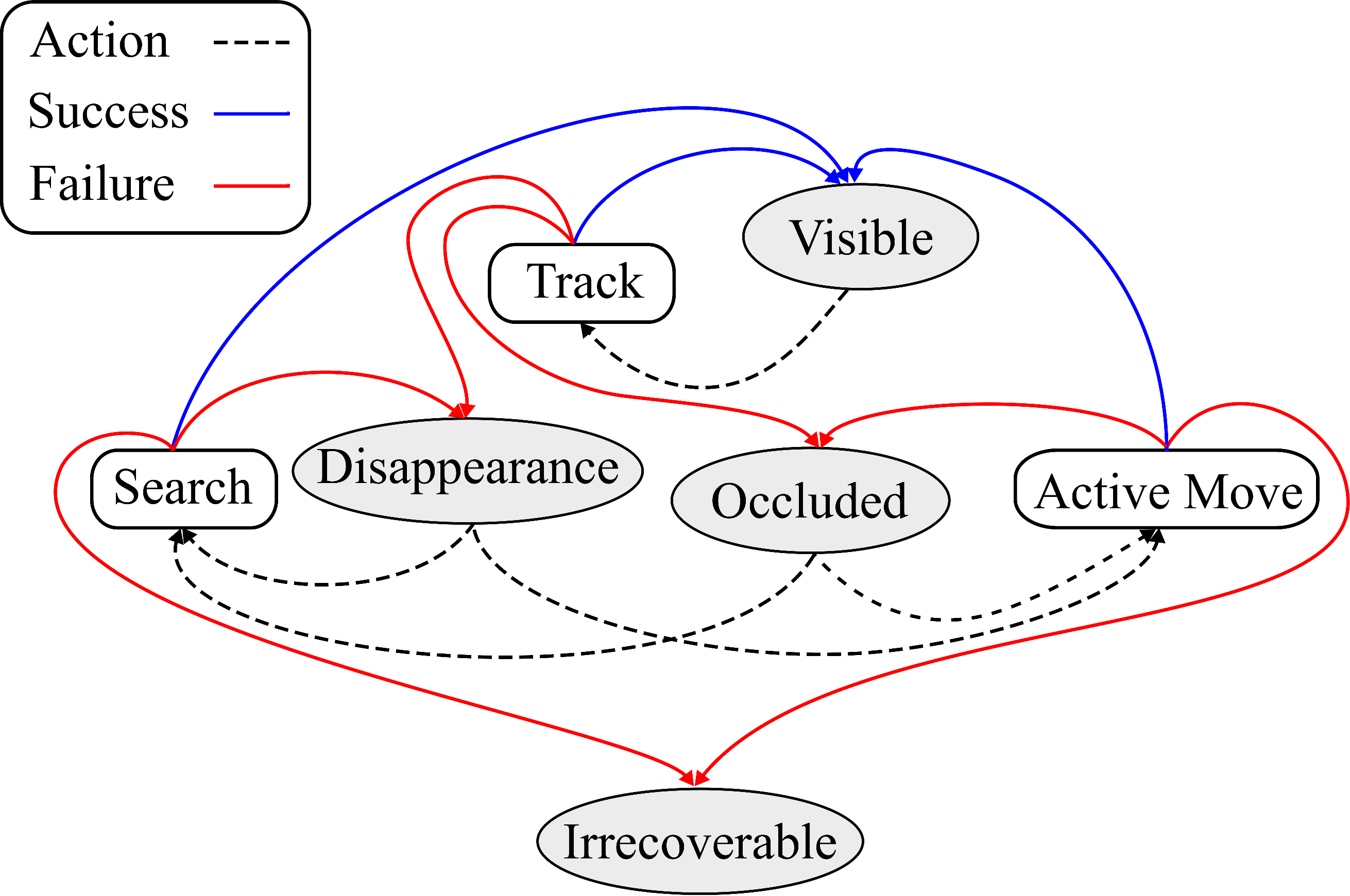}}
       \caption{Transition diagram between context states. The actions are Search, Track, and Active Move. The context states are shaded in gray. When an action doesn't result on finding the target, i.e. Visible state, we consider that the action is in the Fail state. Otherwise it's a Success.}
       \label{Figure:POMDP-STATES}
\end{figure}

\subsubsection{Actions}
Our action set is defined as \textit{A} $=\{$ \textit{Track, Search, Active Move}$\}$. The \textit{Track} action commands the robot to track targets by either panning its head or navigating towards the target. \textit{Active Move} prompts the robot to change its location to better track a visible target or to overcome Occlusion situations where the target is believed to be behind another object. Lastly \textit{Search} prompts the robot to explore its environment in search of humans by turning its head. This action corresponds to the hypothesis that when an object is missing it might be around a human therefore finding humans might reveal the target. In addition, we search for humans instead of directly searching for the target, since people are larger and easier to spot from afar.
   
\subsubsection{Transitions}
\label{pomdp_transition}
We define a transition model to estimate the next context state based on the available actions, i.e. $p(c_k |c_{k-1}) $ as defined in Equation \eqref{Equation: Belief Updates}. We remark that context updates are performed at asynchronous time frames since actions take more than one time step to complete. Accordingly, we define transitions as: 
\begin{equation}
     p(c_l |c_{l-1}) = 
     \begin{cases}
         I & \text{if }  a \; \text{is active}\\
        b_{l-1}(s|a_{l-1}) &  \text{if }  a \; \text{is complete}, \end{cases}
  \end{equation}
where $l$ is a distinct time frame from $k$ since transitions among context states are determined based on the execution time of high-level actions. Transitions between context states are described in Fig. \ref{Figure:POMDP-STATES}.

In order to sample possible scenarios from the current belief using an online POMDP solver such as DESPOT or POMCP \citep{silver2010monte}, we define a deterministic simulative model function $g: S \times A \times R \mapsto S \times Z $,  where a random number $\phi$ is distributed uniformly over [0,1]. Then, $(s',z')=g(s,a,\phi)$ is distributed according to $p(s',z'|s,a)= T(s,a,s')O(s'a,z')$. In our study, the observation function $O(s'a,z')$ varies according to the combination of the sensed robot and target positions, and the occlusion status.

\subsubsection{Observations}
To estimate hidden context states, we rely on four features. The most useful feature is whether a target observation exists or not. If $z_k$ is non-empty, the context state might be \textit{visible} with high probability. Thus, the first feature variable is defined as $\theta_{Target}=1$ if $z_k$ exists, otherwise, $\theta_{Target}=0$.
\begin{figure*}[t]
 \centering 
       \includegraphics[width=1.0\linewidth]{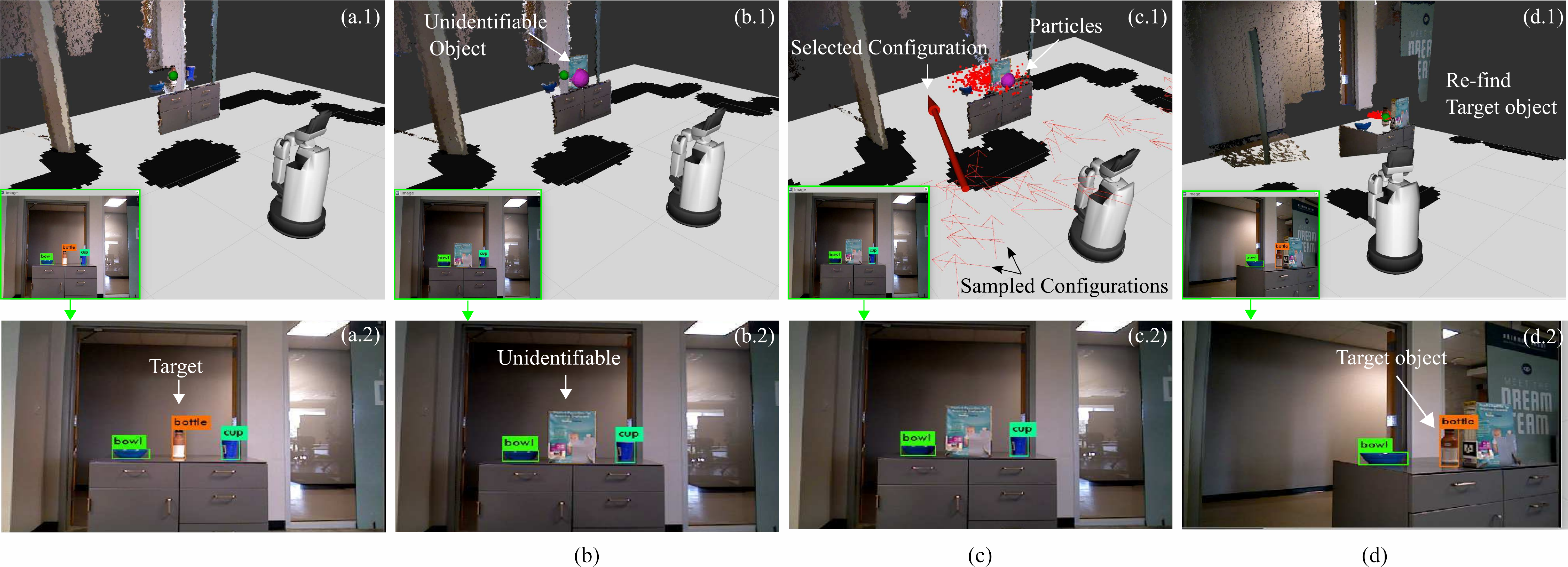}
        \caption{
               Example resolving an occlusion event and leading to re-locating the target. (a) The robot is tracking a target object (a bottle) using its cameras. (b) an occlusion suddenly occurs (e.g. a person has placed a box in front of the bottle); an occluding marker (a purple ball in the figure) is placed in the computer visualization window in front of the target. (c) Using our proposed active sensing  method, the robot chooses its next configuration (shown as a big red arrow) in order to maximize its information gain, which is approximated by particles. (d) For this example, the robot finally succeeds to re-locate the target behind the occluding object. 
        }
        \label{Figure:rviz_view} 
\end{figure*}

The second feature we consider reflects the probability of occlusions to occur, and can be computed by using existing object recognition algorithms. Once an occlusion occurs, the object being tracked will be occluded by another object (both of them represented by bounding boxes traced by the object recognition software). In the case that objects causing occlusions are semantically identifiable, an overlap ratio is defined based on the two bounding boxes (the object being tracked, $i$, and the newly occluding object, $j$) as
\begin{equation}\label{Equation:OverlapRatio}
   \text{OR}_k^{ij}= \frac{ {BB}^j_k \cap {BB}^{i}_{k-1} }{{BB}^i_{k-1}},
\end{equation}
where $BB^i_k$ denotes the bounding box for object $i$ at time index $k$. We define a feature $\theta_{OR}^i \in \{0,1\}$ as the rule $(OR^{ij} > \lambda_{OR}) \cap  (z=\emptyset)$ where $\lambda_{OR}$ is a predefined threshold and $z=\emptyset$ denotes the absence of a target.  
In the case that the occluding object is unidentifiable, the depth variance can also be used to detect the occlusion. Note that the depth within the bounding box will become smaller when a new object starts occluding the previous one. Therefore, features affected by depth variation can be formulated as  
\begin{equation}
 \label{Equation:depth information}
   \Delta{Z}^i_k = \bar{Z}^i_{k-\Delta T:k} - Z^j_k,
\end{equation}
where $\bar{Z}^i_{k-\Delta T:k} = \frac{1}{\Delta T} \sum_{k-\Delta T}^{k} Z^i$ denotes the average depth for time period $\Delta T$ within $BB^i$, and $Z^j_k$ stands for the newly detected depth value (i.e. the occluding object) for the same region. Thus, we define a feature variable for depth information as follows 
\begin{equation}
    \theta^i_{Depth} =  \begin{cases} 
        1 &  \left( \Delta{Z}^i_t > \lambda_{depth} \: \cap \: z=\emptyset \right) \\
        0 & \text{otherwise.} \qquad  \qquad  \qquad 
        \end{cases}. 
\end{equation}

Lastly, the loss of a target can also be inferred based on the premise that the main cause of its disappearance is a human taking it away. This assumption seemingly makes sense since most common objects cannot change their location by themselves. Accordingly, we define a new feature associate with the presence of a human as, $\theta_{Human}=1$ if a human is present. In practice, we detect humans by using the object recognition algorithm \citep{redmon2016you}. 

Consequently, the observation model is expressed using the feature vector 
\begin{equation}
 \Theta:=(\theta_{Target}, \theta_{OR}, \theta_{Depth}, \theta_{Human}).
 \label{eq:feature-vector}
\end{equation}

By detecting features, we estimate the context state using the likelihood distribution $p(o|s,a) \approx p(\Theta|s)=\Pi_{i=1}^4 p(\theta_i |s)$ which relies on the assumption that each feature is conditionally independent.

 We utilize the sensor model from subsection \ref{sensor_model} which depends on whether the target is within the FOV or it is not. In addition, to reflect the decrease in recognition accuracy as the distance between the sensor position and target position $d$ increases, the depth coefficient can be computed as 
\begin{equation}
\xi(d) = \frac{1}{1+\exp{\beta(d-d_{max})}},
\end{equation}
where $\beta$ denotes the slope of the reduction over depth and $d_{max}$ is the maximum distance that targets can be recognized. This depth coefficient is multiplied by the basic feature recognition accuracy parameters $P_d$ and $P_e$ in the sensor model, Equation (13), for use in the observation function of DESPOT.

 \subsubsection{Rewards}
We assign positive rewards only when the context state is \textit{visible} otherwise we do not give rewards. As such, we compute the expected reward $\rho(\cdot)$ using the belief state $b(s)$ and the reward function $r(s,a)$ as 
\begin{align}
&\rho(b(s), a) = \sum_{s}b(s)r(s,a),  \\
&r(s,a) = \left \{\begin{array}{lr}
         \psi* IG(q_k)+R & \qquad \text{if } \;  s \;\;  is \;\; \text{Visible}\\
        \psi* IG(q_k) &  \qquad  \text{otherwise} \qquad  \end{array} \right \},
\end{align}
where $\psi$ denotes the coefficient for normalizing information gain and $R$ is the positive reward for successful re-targeting. In this study, we set the discounting factor, $\gamma$ as 0.9.

\subsection{Context Models}
As indicated in Equation \eqref{Equation:GMM}, each context state has its own prediction model regarding the target's location given prior knowledge.

\subsubsection{Visible Context}
In this case, we predict the target's position from the last target state $x_{k-1}$ using a Gaussian distribution, i.e.
\begin{equation}
  p(x_k |c=\textit{visible}) \approx \mathcal{N}(x_k;x_{k-1}+v_{k-1} \Delta t,\sigma_x^2)
\end{equation}
where $\sigma_x^2$ is the sensor noise.

\subsubsection{Occluded Context}
In this case, we estimate the target's position from the fact that is behind the occluding object's position, $x_{occ}$, i.e. 
\begin{equation}
    p(x_k |c=\textit{Occluded}) \approx \mathcal{N}(x_k;x_{occ}+\delta_{\text{offset}},\sigma_{occ}^2),
\end{equation}
where $\delta_{offset}$ is an approximate distance value such as the length of the bounding box describing the occluding object.

\subsubsection{Disappearance Context}
In this state, the position of the target is inferred from knowledge about nearby people. If no one is visible in the FOV, the robot will begin to look for nearby people by rotating its base and head using the \textit{search} action. Once a person has been detected, a prediction model will generate particles based on the following model,
\begin{equation}
  p(x_k|c=\textit{Disappearance}) \approx \mathcal{N}(x_k;x_{human},\sigma_{human}^2).
 \end{equation}
 
 \begin{figure}[t]
 \centering 
       \includegraphics[width=1.0\linewidth]{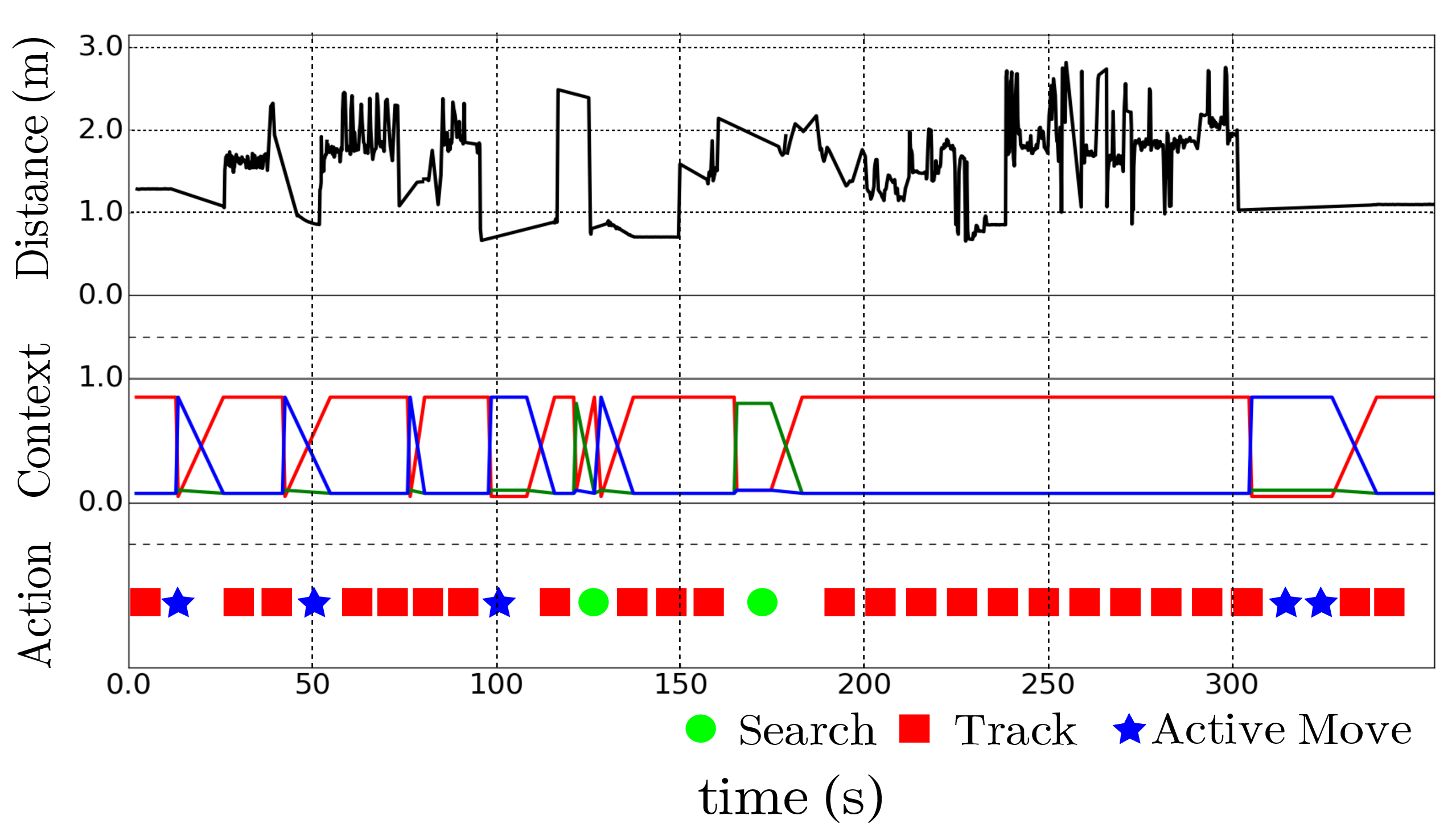}
        \caption{Time analysis for an example scenario of active object tracking. The first row shows the estimated distance between the robot and the target object. The second row shows the estimated context state (red line: \textit{visible}, blue line: \textit{Occluded}, and green line: \textit{Disappearance}). The third row provides the chosen actions at each time frame.}
        \label{Figure:scenario_analysis} 
\end{figure}

\section{Experimental Results}
\subsection{Robot Description}
To validate our approach experimentally, we use the Toyota Human Support Robot (HSR), a mobile manipulator equipped with an omnidirectional mobile base and a panning and tilting head. A depth camera (Xtion, Asus) is located on top of the robot's head to obtain an RGB-D stream, sampled with 30fps. The camera parameters for FOV are 58$\degree$ H, 45$\degree$ V, 70$\degree$ D, which stand for the angle for horizontal, vertical, and diagonal, respectively. A laser range scanner, Hokuyo, is installed at the front bottom of the base in order to detect static and dynamic obstacles. The HSR uses two different computers, one Intel Core i7, 4th Gen with 16GB RAM is used for basic navigation functions of the robot and another one, an Alienware Intel Core i7-7820HK, GTX 1080 laptop, is used for running object detction algorithm. Interprocess communications are handled with ROS. All the skills for robots are written as ROS actions so that initiation and termination of actions can be easily managed from a high level planner. The testing facility is the UT Austin's Human Centered Robotics Lab. For object detection we have used YOLOv3 \cite{redmon2018yolov3}. It is regarded as one of the state-of-the-art methods, because of its accuracy and speed. Retinanet \cite{lin2017focal} and SSD \cite{liu2016ssd} are also viable algorithms for our research, but YOLOv3 has the advantage of fast inference times. During our experimentation, it was assumed that if an object exists within the camera's FOV, it will be detected reliably through YOLOv3. Based on this reliability, the value of the true positive probability of the sensor model, $p_d$ was set to 0.95. Our focus thereafter has been on inference of occlusions or the response of the planner when objects go missing.

\begin{figure*}
 \centering 
       \includegraphics[width=1.0\linewidth]{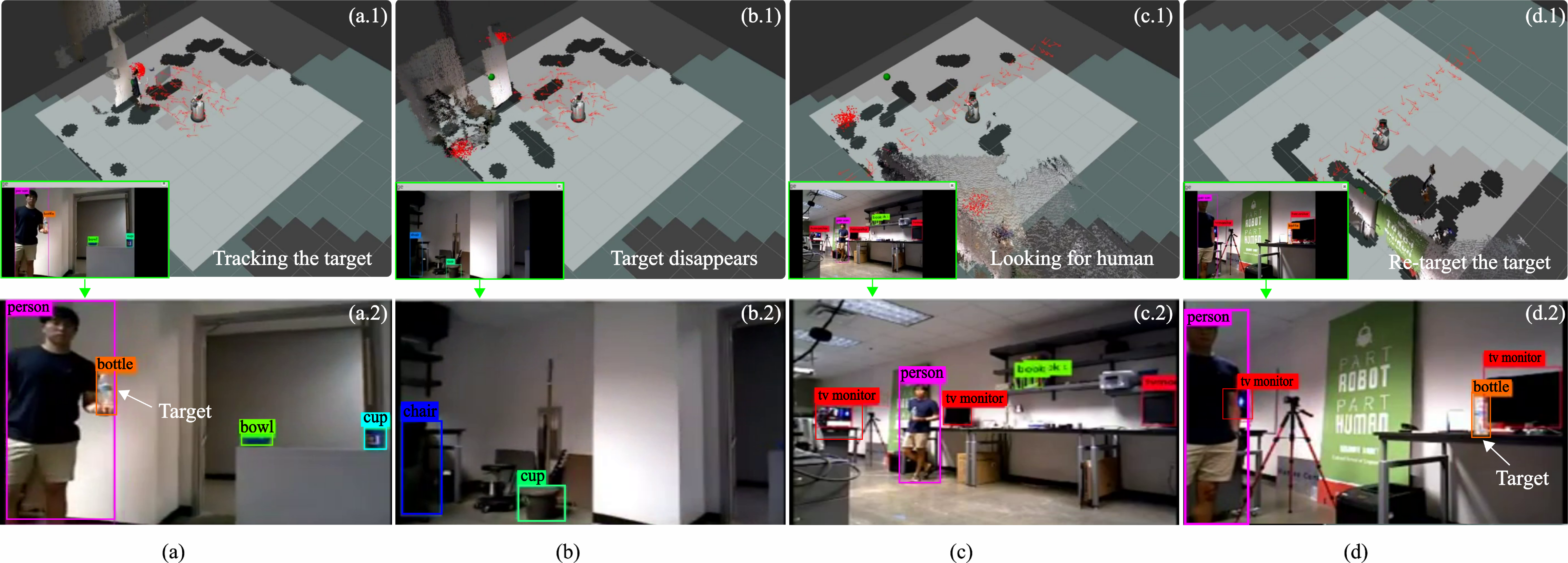}
        \caption{
               Demo showcasing finding a target when it suddenly disappears from the robot's FOV and is not believed to be occluded by another object. (a) Initially, the robot tracks the target (a bottle) using its depth cameras. (b) The target suddenly disappears from its FOV; here it is believed the target has disappeared because of the information returned from the observation model described earlier. (c) The robot moves to find a person around its neighborhood. If a person is detected, the robot navigates to that person and attempts to locate the target nearby. (d) In our demo, the robot succeeds in re-locating the object since it was placed next to the person.
        }
        \label{Figure:rviz_view_disappearance} 
\end{figure*}

\subsection{Scenarios}
We deal with three possible situations: 1) occlusions occur due to the interference of another object, 2) objects disappear when they are taken away by people, and 3) objects temporarily move outside of the FOV but can be quickly found by employing active tracking. Fig. \ref{Figure:rviz_view} shows that the robot is able to track and resolve occlusion situations through searching for new configurations and successfully re-locating the target. In addition, the examples demonstrating the second and third cases above are shown in Fig. \ref{Figure:rviz_view_disappearance}. Details can be found in Fig. \ref{Figure:scenario_analysis}. This figure provides the evolution of the estimated context states and the corresponding high-level actions taken over time based on context states.  

\subsection{Performance Evaluation}
We propose four criteria to evaluate performance: Success Ratio (SR), Tracking Ratio (TR), Average Restoring Time (ART), and Failure Time (FT). Each criterion is evaluated over 20 trials until the robot fails to track the target. The statistical results are shown in Table \ref{table1}.

\subsubsection{Success Ratio (SR)}
\begin{figure}[t]
 \centering 
      \includegraphics[width=0.78\linewidth]{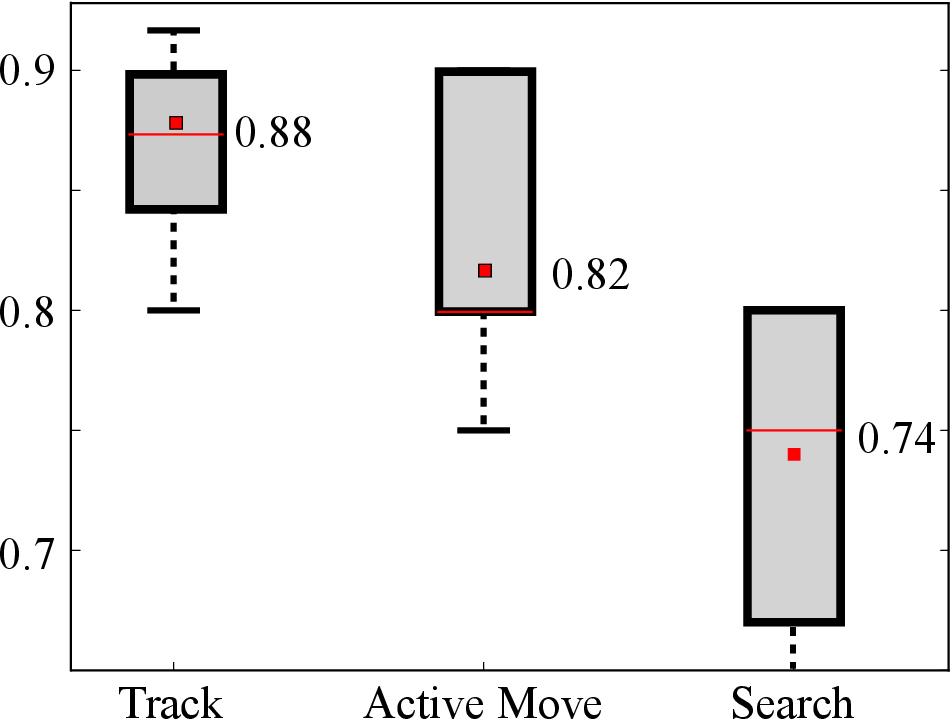}
        \caption{ Success ratio for various active sensing scenarios. The rates represent the ratio of success to track or find a target after action execution depending on believed context states. \textit{Track} behaviors have the highest success rates, 0.88, followed by \textit{Active Move}, while \textit{search} has the highest variance and the lowest success rates, 0.74.}
        \label{Figure:boxplot of each success ratio} 
\end{figure}

Success ratio given an action represents the number of successful target re-locations versus the total number of trials. In our experiments, we achieved an overall success rate of 0.82 with standard deviation equal to 0.097 despite the conditions being highly dynamic (i.e. the target is moving or occluded or has suddenly disappear). The success ratio case is shown for each context and action being taken in Fig. \ref{Figure:boxplot of each success ratio}.  

\subsubsection{Tracking Ratio (TR)}
This ratio can be regarded as keeping the target in the FOV. It is calculated using the amount of time the target is believed to be in the visible state versus the total time an experiment lasts. We achieved an average TR value of 0.7 for the demos above, i.e. a mixture of experiments where we repeatedly occluded the target, or a person moved it away to nearby locations outside of the robot's FOV, or a person rapidly moved the target around the robot.

\subsubsection{Average Restoring Time (ART)}
This time is calculated as the differential of time between loosing sight of a target until it is re-located. It varies depending on the type of action. For example, the average restoring time for \textit{Track} was 2 seconds while that for \textit{ActiveMove} was 12.15 seconds with a standard deviation of 3.95. Lastly, the ART for \textit{Search} was 16.5 seconds with standard deviation of 7.8. 
 
\subsubsection{Failure Time (FT)}
Lastly, we measure the amount of time it takes the robot to fail tracking or searching targets.
If the target is not found within 1 minute, the robot will go into failure mode, i.e. the context state is irrecoverable. The experiment for this measurement is done based on an arbitrary mixture of context states performed by moving the target around or occluding it. The overall FT for our action sequence is 232 seconds with standard deviation of 44.2 seconds. This indicates that the proposed algorithm performs successful target tracking and searching tasks for approximately 4 minutes before becoming irrecoverable due to arbitrary user manipulations.
  
 \begin{table}[t]
\caption{Four Criteria results}
\centering
\begin{tabular}{c||c|c}
\hline
\hline
Criterion & Mean & Standard Deviation \\
\hline
SR  & 0.82  & 0.097  \\
TR & 0.71 &  0.096 \\
ART (s) & 10.22  & 7.9 \\
FT (s) &  232 &  44.2  \\ 
\hline
\hline
\end{tabular}
\label{table1}
\end{table}
  
\section{Concluding Remarks}
This paper addresses active target tracking and searching capabilities using mobile robots. Experimental results are performed using multiple  scene trials. To this point, we employ information-theoretic costs for active search. Integrating a DBN model, particle filtering and POMDP planning, we are able not only to infer target locations from context information in a probabilistic manner, but also to define cost functions effectively. Obtaining the desirable control inputs for gathering information allows our robot to have better tracking and search capabilities under several dynamic conditions, such as occlusions and the sudden disappearance of a target. 

We highlight the following achievements. First, our methods are scalable and versatile via the proposed hybrid state estimator, i.e. the particle filter plus POMDP based on context states. To this point, the action sequence shown in Fig. \ref{Figure:scenario_analysis} effectively demonstrates the robot's capability to actively find occluded or missing targets under various dynamic conditions. Second, as shown in Fig. \ref{Figure:boxplot of each success ratio}, our active sensing algorithm, on average, re-locates targets with high success rates. Finally, we have verified robustness and efficiency of our methods using the proposed criteria, SR, TR, ART, and FT. In essence, we verified that the robot can perform practical tracking and search tasks while operating in dynamic environments. 

This paper has been focused on the estimation of objects and the control of the robot. The experiment presented was designed as a concept validation. In the future we will deploy the framework in many different scenarios to obtain statistical data for in-depth analysis. We will also be on the lookout for benchmark data-sets for "active object tracking" since to date there don't exist. Furthermore, better probabilistic models are required to model realistic target search situations. For example, our current method assumes that direct transitions between occlusion and disappearance states do not happen. Furthermore, to deal with more realistic situations, a memory-based approach that uses historical data should be used to predict context. Recurrent Neural Networks might be a good solution for future work. Lastly, there exists a room for extending our proposed work to multi object tracking cases.


Overall, this work demonstrates a prototype of robust autonomous active target tracking performed using a mobile robot in a practical setup.

\begin{appendices}

\section{Parameter Selection}
 
 
For particle filtering during experimentation, there are various parameters to be determined, such as the total number of particles, re-sampling conditions (effective sample size) or type of method (residual method), motion model noise, or the sensor model. Here are the parameter list that we empirically determined for our study. 

\begin{table}[]

\caption{Parameters for Particle Filter}
\centering
\begin{tabular}{|c||c|c|}
\hline
Parameter & Description  & Value  \\
\hline
$N_{total}$  & Total number of particles  & 2000 \\
\hline
$N_{ESS}$ & Effective Sample Size &  1000   \\
\hline
$p_d$ & True positive probability of detection  & 0.95\\
\hline
$\Sigma$ &  Variance of detection    & 0.25\\ 
\hline
\end{tabular}
\label{table1}

\end{table}

\begin{table}[]

\caption{Parameters for POMDP}
\centering
\begin{tabular}{|c||c|c|}
\hline
Parameter & Description  & Value  \\
\hline
$\psi$  & Reward Coefficient & 0.1\\
\hline
$R$ & Positive Reward for finding the target &  10.0   \\
\hline
$\gamma$ & Discounting factor  & 0.9\\
\hline
$ d_{max}$&  Maximum distance for target detection   & 10.0\\ 
\hline
$ \delta_{\text{offset}}$ &  Distance to target object from occluder  & 0.35\\ 
\hline
$ \sigma_x^2$ &  Detection noise  & 0.05\\ 
\hline
$ \sigma_{occ}^2$ &  Variance of occluded target location   & 0.15\\ 
\hline
$ \sigma_{human}^2$ & Variance of human location    & 0.5\\ 
\hline
\end{tabular}
\label{table1}
\end{table}


\end{appendices}

\begin{acknowledgements}
The authors would like to thank the members of the Human-Centered Robotics Laboratory at The University of Texas at Austin for their great help and support. This work was supported by ONR Grant $\#$N000141512507. The research was also partially sponsored by the Army Research Office and was accomplished under Cooperative Agreement Number W911NF-19-2-0333. The views and conclusions contained in this document are those of the authors and should not be interpreted as representing the official policies, either expressed or implied, of the Army Research Office or the U.S. Government. The U.S. Government is authorized to reproduce and distribute reprints for Government purposes notwithstanding any copyright notation herein. 
\end{acknowledgements}

\bibliographystyle{spbasic}
\bibliography{reference}

\end{document}